\documentclass{article}

\PassOptionsToPackage{numbers}{natbib}
\usepackage[preprint]{neurips_2024}

\usepackage{graphicx}
\usepackage{amsmath}
\usepackage{amssymb}
\usepackage{booktabs}
\usepackage{multirow}
\usepackage{mathrsfs}
\usepackage{subcaption}
\usepackage{pifont}
\usepackage[table]{xcolor}

\definecolor{cvprblue}{rgb}{0.21,0.49,0.74}
\usepackage[pagebackref,breaklinks,colorlinks,citecolor=cvprblue]{hyperref}
\usepackage{enumitem}
\usepackage{wrapfig}
\usepackage{adjustbox}

\definecolor{rowunit}{RGB}{128,128,255}
\definecolor{evaunit01green}{RGB}{54,125,189}

\usepackage{tabularx}
\usepackage[capitalize]{cleveref}
\crefname{section}{Sec.}{Secs.}
\Crefname{section}{Section}{Sections}
\Crefname{table}{Table}{Tables}
\crefname{table}{Tab.}{Tabs.}

\usepackage[utf8]{inputenc} % allow utf-8 input
\usepackage[T1]{fontenc}    % use 8-bit T1 fonts
\usepackage{hyperref}       % hyperlinks
\usepackage{url}            % simple URL typesetting
\usepackage{booktabs}       % professional-quality tables
\usepackage{amsfonts}       % blackboard math symbols
\usepackage{nicefrac}       % compact symbols for 1/2, etc.
\usepackage{microtype}      % microtypography
\usepackage{xcolor}         % colors
\usepackage{listings}
\usepackage{makecell}
\definecolor{evaunit02red}{RGB}{195,056,040}
\newcommand{\evared}[1]{\textcolor{evaunit02red}{#1}}
\newcommand{\dplus}[1]{\fontsize{7pt}{0.1em}\selectfont (\textbf{\evared{#1}})}

\newcommand{\logotitle}{%
  \begin{minipage}{0.1\textwidth}
    \includegraphics[width=1.2cm]{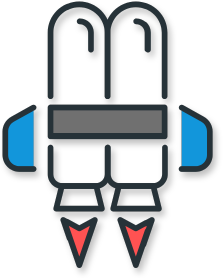}%
  \end{minipage}%
  \begin{minipage}{0.85\textwidth}
    \centering
    \textbf{MoE Jetpack: From Dense Checkpoints to } \\
    \textbf{Adaptive Mixture of Experts for Vision Tasks}%
  \end{minipage}%
}

\title{\logotitle{}}

\author{%
Xingkui Zhu$^{*}$\quad Yiran Guan\thanks{\footnotesize{Equal contribution. $\dagger$ Corresponding author.}} \quad Dingkang Liang\quad Yuchao Chen\quad \\
\textbf{Yuliang Liu}\quad \textbf{Xiang Bai}$^{\dagger}$\\
 Huazhong University of Science and Technology\\
 \texttt{\{adlith, yiranguan, xbai\}@hust.edu.cn}
}

\begin{document}

\lstset{
  basicstyle=\small\ttfamily,
  columns=fullflexible,
  keepspaces=true,
  breaklines=true,
  breakatwhitespace=true,
  language=Python,
  keywordstyle=\color{blue},
  stringstyle=\color{purple},
  commentstyle=\color{teal},
  showstringspaces=false,
}

\maketitle

\begin{abstract}
The sparsely activated mixture of experts (MoE) model presents a promising alternative to traditional densely activated (dense) models, enhancing both quality and computational efficiency. However, training MoE models from scratch demands extensive data and computational resources. Moreover, public repositories like timm mainly provide pre-trained dense checkpoints, lacking similar resources for MoE models, hindering their adoption. To bridge this gap, we introduce MoE Jetpack, an effective method for fine-tuning dense checkpoints into MoE models. MoE Jetpack incorporates two key techniques: (1) \textit{checkpoint recycling}, which repurposes dense checkpoints as initial weights for MoE models, thereby accelerating convergence, enhancing accuracy, and alleviating the computational burden of pre-training; (2) \textit{hyperspherical adaptive MoE (SpheroMoE) layer}, which optimizes the MoE architecture for better integration of dense checkpoints, enhancing fine-tuning performance.
Our experiments on vision tasks demonstrate that MoE Jetpack significantly improves convergence speed and accuracy when fine-tuning dense checkpoints into MoE models. 
Our code will be publicly available at \url{https://github.com/Adlith/MoE-Jetpack}.

\end{abstract}

\vspace{-2mm}
\section{Introduction}
\label{Introduction}
Increased scale is one of the key factors boosting performance in deep learning~\cite{alexnet,resnet,vitg}. However, as models expand in size, their computational demands surge, resulting in considerable slowdowns during both training and inference phases. A promising approach that decouples model size from computational costs is the \textit{sparsely activated mixture of experts} (MoE)~\cite{fedus2022switch,riquelme2021scaling,softmoe}. Unlike \textit{densely activated models} (referred to as \textit{dense models} hereafter)~\cite{VIT,liu2022convnet} apply the full network parameters to all inputs, MoE dynamically activates different pieces of the model for distinct input tokens. This allows for model scaling without substantially increasing the FLOPs\footnote[1]{FLOPs means the floating point operations per second. The vanilla design of MoE does not inherently provide runtime advantages and requires additional parallelization strategies~\cite{fan2022m3vit, rajbhandari2022deepspeed} for acceleration. In our implementation, we offer an effective matrix multiplication method for parallelization, detailed in Appendix~\ref{pseudo code}.}, thereby maintaining training and inference speeds during model upscaling. Recent advancements have seen successful implementations of MoE across various domains~\cite{mixtralmoe,deepseekmoe,moellava}.

Despite their potential, MoE models face significant adoption challenges primarily due to the lack of pre-trained models. Unlike dense models, which benefit from a rich repository of pre-trained models accessible through communities like Hugging Face~\cite{wolf2019huggingface} ($\sim$81k models) and Timm~\cite{rw2019timm} ($\sim$800 models), most MoE models must be trained from scratch using randomly initialized weights. The absence of pre-trained weights necessitates substantial GPU hours and extensive data for training Mixture of Experts (MoE) models, thereby restricting MoE research to a limited number of research teams.
Consequently, our research aims to reduce the training time and data requirements for MoE models by leveraging the pre-trained knowledge from dense checkpoints. \textit{We will specifically investigate whether utilizing dense checkpoints can enhance the accuracy and convergence speed of MoE models during fine-tuning.}

\begin{figure}[t]
\vspace{-5mm}
\begin{center}
   \includegraphics[width=1.0\linewidth]{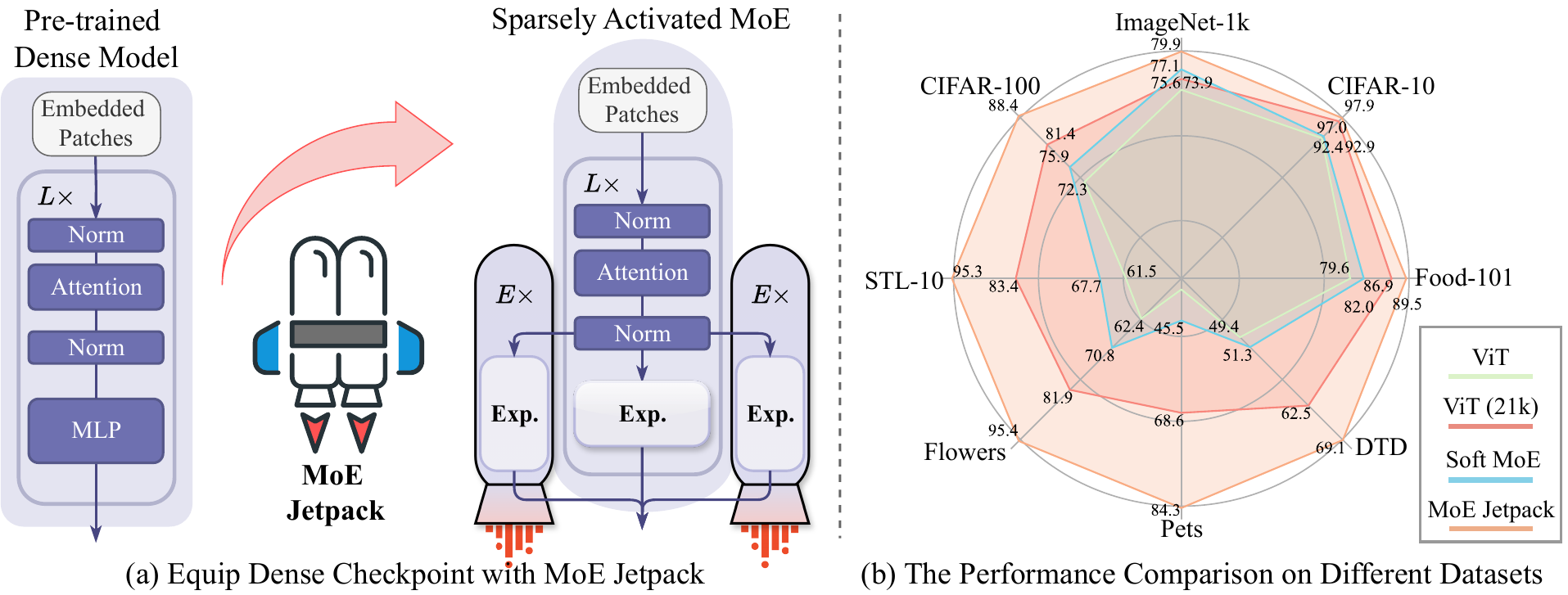}
\end{center}
% \vspace{-5pt}
   \caption{
    (a) Our MoE Jetpack converts pre-trained dense models into MoE models, enhancing convergence and performance while maintaining equivalent FLOPs. Here, \textbf{Exp.} represents individual experts, \(E\) denotes the number of experts, and \(L\) indicates the total number of layers.
    (b) Performance comparison of ViT trained from scratch, pre-trained ViT, Soft MoE~\cite{softmoe} trained from scratch, and MoE Jetpack across various datasets. MoE Jetpack shows significant performance improvements.
   }
\label{fig:MoE Jetpack}
\vspace{-3mm}
\end{figure}

In this paper, we propose MoE Jetpack, a new approach for fine-tuning pre-trained dense checkpoints into MoE models. As illustrated in Fig.~\ref{fig:MoE Jetpack}(a), MoE Jetpack leverages the sunk cost of dense pre-training to enhance MoE model performance and accelerate convergence. It comprises two key techniques. 
The first is \textbf{checkpoint recycling}, which initializes MoE models using dense checkpoints. Unlike sparse upcycling~\cite{komatsuzaki2022sparseUpcycling}, which merely copies the Multilayer Perceptron (MLP) to construct experts, checkpoint recycling leverages various dense checkpoints and multiple weight selection methods. This approach provides greater flexibility and results in superior MoE initialization weights.
The second technique is the \textbf{hyperspherical adaptive MoE (SpheroMoE) layer}, which presents an optimized MoE architecture for seamless integration of dense checkpoints and enhanced fine-tuning performance. Existing MoE architectures, such as Switch Transformers~\cite{fedus2022switch} and Soft MoE~\cite{softmoe}, are not designed to leverage pre-existing dense checkpoints, which may lead to optimization and over-specialization challenges during fine-tuning. The SpheroMoE layer mitigates these challenges by normalized token mixing, expert regularization, and adaptive dual-path.

By equipping dense checkpoints with MoE Jetpack, the fine-tuned MoE models achieve significantly higher accuracy than those trained from scratch while maintaining the same FLOPs as the original dense models. Comprehensive evaluations of MoE Jetpack across various image classification datasets of differing scales demonstrate its effectiveness. As shown in Fig.~\ref{fig:MoE Jetpack}(b), MoE Jetpack leverages existing dense checkpoints to enhance fine-tuning convergence speed and performance, maintaining efficiency comparable to the original dense models. In summary, our contributions are as follows:
\begin{itemize}[leftmargin=*]
\item We introduce \textit{checkpoint recycling}, which pioneers the selection of dense checkpoints to initialize MoE experts, enhancing initialization flexibility, diversifying experts, and eliminating the computational burden of MoE pre-training.
\item We develop the \textit{spheroMoE layer}, optimized for fine-tuning dense checkpoints into MoE architectures, alleviating optimization challenges, and preventing the over-specialization of experts.
\end{itemize}

\section{Background}
\label{Background}
In this section, we recap the main components used in MoE Jetpack: the sparsely activated mixture of expert (MoE) architectures and different routing mechanisms in MoE. Existing MoE models are typically derived from dense Vision Transformer (ViT) architectures by replacing certain multilayer perceptron (MLP) layers with MoE layers. Each MoE layer comprises a $\text{Router}(x; \theta_{gate})$ and several "experts", each parameterized as $\text{MLP}(\cdot; \theta_i)$. 
MoE models typically use similar experts, with differences primarily in their routing mechanisms. Several routing algorithms have been developed, including Top-K~\cite{shazeer2016outrageously}, BASE and Sinkhorn-BASE layers~\cite{lewis2021base,clark2022unified}, Hash layers~\cite{roller2021hash}, Expert Choice routing~\cite{zhou2022mixture}, and soft routing~\cite{softmoe}.
The most commonly employed routing mechanism is the Top-K, which effectively reduces computational overhead by sparsely activating only the top-K experts relevant to the input token. The routing decision is formulated as:
\begin{equation}
\text{Router}(x; \theta_{gate}) = \text{Top-K} \left( \text{softmax}(\text{MLP}(x; \theta_{gate})) \right),
\end{equation}
% the output is then computed by aggregating the contributions of these experts:
\begin{equation}
y = x + \sum_{i \in E} \text{Router}(x; \theta_{gate}) \cdot \text{Expert}(x; \theta_i),
\end{equation}
where $\theta$ denotes the weights, \(E\) is the set of activated experts, and \(|E| = K\).
However, Top-K routing faces challenges such as imbalanced expert utilization, token dropping, and scalability issues. 

Enhanced mechanisms like Soft MoE~\cite{softmoe} address these issues and serve as our baseline. 
The Soft MoE routing algorithm processes input tokens \(\mathbf{X} \in \mathbb{R}^{m \times d}\), where \(m\) represents the number of tokens, and \(d\) is their dimensionality. It uses learnable parameters \(\Phi \in \mathbb{R}^{d \times (e \cdot s)}\) to reconfigure these tokens into \(e \times s\) slots. The transformed input slots \(\tilde{\mathbf{X}} \in \mathbb{R}^{ (e \cdot s) \times d}\) are combinations of the input tokens: \(\tilde{\mathbf{X}} = \text{softmax}(\mathbf{X}\Phi)^\top \mathbf{X}\). Each MoE layer includes \(e\) expert functions \(\{f_i : \mathbb{R}^d \rightarrow \mathbb{R}^d\}_{i=1}^e\), with each expert handling \(s\) slots.
The intermediate output \(\tilde{\mathbf{Y}} \in \mathbb{R}^{(e \cdot s) \times d}\) is obtained by applying the expert functions to the transformed slots: \(\tilde{\mathbf{Y}}_{i,j} = f_i(\tilde{\mathbf{X}}_{i,j})\) for \(i \in \{1, \ldots, e\}\) and \(j \in \{1, \ldots, s\}\).
The output tokens \(\mathbf{Y} \in \mathbb{R}^{m \times d}\) are generated by reassembling the outputs of the experts: \(\mathbf{Y} = \text{softmax}(\mathbf{X}\Phi)\tilde{\mathbf{Y}}\).

\section{MoE Jetpack}
\label{Method}
In this section, we present the overarching concept of the MoE Jetpack. It is divided into two phases: initializing MoE models with checkpoint recycling and fine-tuning MoE models using the hyperspherical adaptive MoE (SpheroMoE) layer.

% \vspace{-2mm}
\subsection{Checkpoint Recycling}
\label{Checkpoint Recycling}
% \vspace{-2mm}
Checkpoint recycling is a foundational phase in the MoE Jetpack framework, transforming pre-trained dense model checkpoints (\textbf{predecessors}) into high-quality initialization weights for MoE models (\textbf{successors}). This approach ensures efficient utilization of resources invested in predecessors, boosting the performance and convergence speed of successors. The recycling procedure involves splitting the predecessors' multilayer perceptrons (MLPs) into multiple experts, ensuring expert diversity and adaptability in expert size to meet varied needs.

To define the process of checkpoint recycling (as illustrated in Fig.~\ref{fig:main_idea}(a)), consider predecessors with $N$ layers $L_i$, a channel dimension of $d$, and a hidden dimension (neuron) of $4d$. We aim to transform these into a successor MoE model $S$ with $N$ layers $L_i'$, a reduced channel dimension $d'$, where $d' \leq d$. Following the Soft MoE~\cite{softmoe}, the successor comprises two segments: a dense part with $N_1$ layers and an MoE part with $N_2$ layers, where $N_1 = N_2 = \frac{N}{2}$. Formally, the successor model is represented as:
\begin{equation}
    S = \left( \{L_i'\}_{i=0}^{\frac{N}{2}-1}, \{L_i'\}_{i=\frac{N}{2}}^{N-1} \right).
\end{equation}
% where $D = \{L_i'\}_{i=0}^{N_1-1}$ and $M = \{L_i'\}_{i=N_1}^{N-1}$.

% For practical illustration, consider extracting $e = 196$ experts from a predecessor with $d = 384$ (as illustrated in Fig.\ref{fig:main_idea}(a)), each expert having a channel dimension of $d' = 192$. These experts are then combined with our routing mechanism to form an MoE layer. 
Inspired by Weight Selection~\cite{xu2024initializing}, our recycling process maintains channel consistency. We explore four primary strategies to guide the recycling of checkpoints:

\begin{figure}[t]
    \vspace{-5mm}
    \centering
    \includegraphics[width=1.0\linewidth]{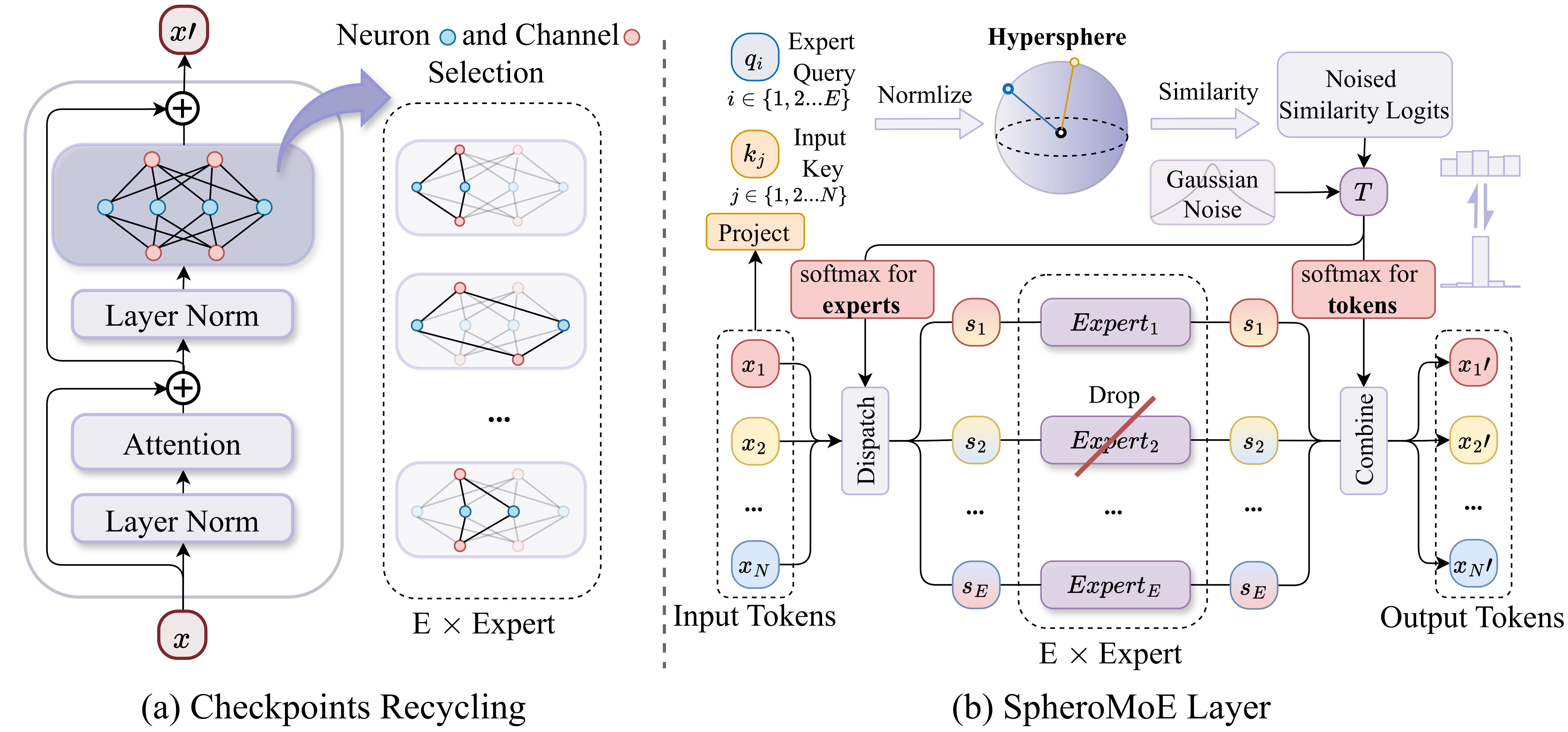}
    % \vspace{2mm}
    \caption{
    (a) Checkpoint Recycling selects neurons and channels from the MLP of pre-trained dense checkpoints using weight sampling methods. This process transforms pre-trained knowledge into multiple experts of any size for initializing MoE models.
    (b) The SpheroMoE layer uses cross-attention to adaptively dispatch input tokens to expert slots. It starts with a randomly initialized query and uses keys and values derived and normalized from the input. The similarity logits between the query and key are calculated in a hyperspherical space, stabilizing the random query. The outputs from the experts are then combined back into the input using the generated similarity logits.
    }
    \label{fig:main_idea}
    \vspace{-3mm}
\end{figure}

% \textbf{Importance-Based Weight Sampling (default)}: Weights are sampled based on their importance, determined by activation values. We pass batches of images through the predecessor and obtain the activation values for each channel and neuron in the MLP layers. For channel selection, we average the activation values across the $N$ layers and select the top-$d'$ channels:
% \begin{equation}
%     A_{c} = \frac{1}{N} \sum_{i=0}^{N-1} A_{L_i}^{c}, \quad \text{Top-}d' = \text{argmax}_{c} (A_{c}), \quad |c| = d'.
% \end{equation}
% For hidden dimensions, we convert the activation values into a probability distribution and sample different experts based on this distribution:
% \begin{equation}
%     h_{\text{successor}} \sim \frac{A_{h}}{\sum_{h' \in H} A_{h'}}, \quad |h_{\text{successor}}| = 4d',
% \end{equation}
% where $A_{h}$ is the activation value of hidden neuron $h$, and $H$ is the set of all hidden neurons. This method ensures that the most important weights are selected for the successor model.

\textbf{Importance-Based Weight Sampling (default)}: Weights are sampled based on their importance, determined by activation values. We pass batches of images through the predecessor model and obtain the activation values for each channel and neuron in the MLP layers. For channel selection, we average the activation values across the \(N\) layers and select the top-\(d'\) channels:
\begin{equation}
    A_{c} = \frac{1}{N} \sum_{i=0}^{N-1} A_{L_i}^{c}, \quad \text{Top-}d' = \text{argmax}_{c} (A_{c}), \quad |c| = d',
\end{equation}
where \(A_{L_i}^{c}\) is the activation value of channel \(c\) in layer \(L_i\), \(A_{c}\) is the averaged activation value of channel \(c\) across \(N\) layers, and \(\text{Top-}d'\) represents the indices of the top-\(d'\) channels with the highest average activation values.

For hidden dimensions, we convert the activation values into a probability distribution and sample different experts based on this distribution:
\begin{equation}
    P(h|H) = \frac{A_{h}}{\sum_{h' \in H} A_{h'}}, \quad h_{\text{successor}} \sim P(h|H), \quad |h_{\text{successor}}| = 4d',
\end{equation}
where \(A_{h}\) is the activation value of hidden neuron \(h\), and \(H\) is the set of all hidden neurons. This method ensures that the most important weights are selected for the successor model.

\textbf{Co-Activation Graph Partitioning}: This strategy groups frequently co-activated neurons into one expert. We construct a co-activation graph by counting the co-activations of neurons in the predecessor for training samples. Each neuron is a vertex in the graph, and edges represent their co-activation frequency. Formally, let $G = (V, E)$ be the co-activation graph, where $V$ represents the neurons and $E$ represents edges with weights indicating co-activation counts. Using the Metis graph partitioning~\cite{karypis1997metis}, we get several subgraphs:
\begin{equation}
    G = \bigcup_{i=1}^{k} G_i, \quad G_i = (V_i, E_i), \quad V_i \cap V_j = \emptyset \text{ for } i \neq j.
\end{equation}
Experts are formed by the combination of sub-graphs. This method leverages the natural grouping of neurons, ensuring each expert captures a specific functional subset of the predecessor model.

\textbf{Uniform Weight Selection}:
Weights are selected uniformly across channels. For a predecessor with channel dimension $d$ and a successor with dimension $d'$, weights are chosen as:
\begin{equation}
W_{\text{successor}}^{(i)} = W_{\text{predecessor}}^{(k)}, \quad k = \left\lfloor \frac{i \cdot d}{d'} \right\rfloor, \quad i \in \{0, \ldots, d'-1\}.
\end{equation}
This method ensures an even distribution of the pre-trained weights across the successor MoE.

\textbf{Random Weight Sampling}:
Weights are randomly selected from the predecessor model. Let $S$ be a random subset of channel indices:
\begin{equation}
S \subseteq {0, \ldots, d-1}, \quad |S| = d'.
\end{equation}
Then, the weights for the successor are chosen as:
\begin{equation}
W_{\text{successor}}^{(i)} = W_{\text{predecessor}}^{(j)}, \quad j \in S, \quad i \in \{0, \ldots, d'-1\}.
\end{equation}
Through the ablation in Sec.~\ref{ablation Checkpoint Recycling}, Importance-Based Weight Sampling is identified as the default method for recycling dense checkpoints to initialize MoE models.

\newpage
\subsection{SpheroMoE Layer}

\begin{wrapfigure}{rh}{0.42\textwidth}
\vspace{-6mm}
\begin{center}
   \includegraphics[width=1.0\linewidth]{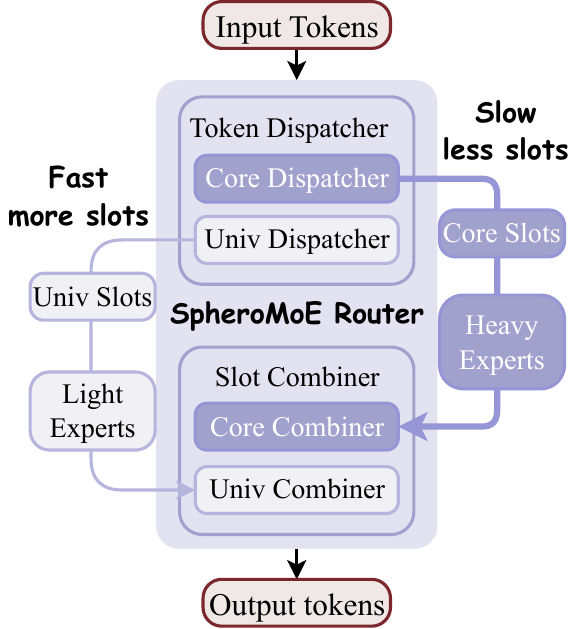}
\end{center}
    \vspace{-2mm}
   \caption{
   The Adaptive Dual-path MoE structure enhances the SpheroMoE Router by adapting it into a dual-branch system, designed to optimize computational efficiency and model performance. This configuration directs high-impact tokens to a core path with fewer but larger experts, while routing less critical tokens to a universal path equipped with a greater number of smaller experts.
   }
\label{fig:dual_path}
\vspace{-6mm}
\end{wrapfigure}

\label{SpheroMoE Layer}
Following the initialization of MoE weights through Checkpoint Recycling, the next step is fine-tuning on downstream datasets. To enhance performance and stability, we designed the hyperspherical adaptive MoE (SpheroMoE) layer (Fig.~\ref{fig:main_idea}(b)), introducing three key improvements: SpheroMoE Routing to alleviate optimization challenges, Expert Regularization to prevent over-specialization, and Adaptive Dual-path MoE (Fig.~\ref{fig:dual_path} for better performance and efficiency. Additionally, the pseudo-code detailing these features' implementation can be found in Appendix~\ref{pseudo code}.

\textbf{SpheroMoE Routing}:
As shown in Fig.~\ref{fig:main_idea}(b), the proposed hyperspherical MoE (SpheroMoE) routing mechanism utilizes cross-attention~\cite{vaswani2017attention} to distribute inputs across experts. Each expert receives an input slot, a weighted average of all input tokens. To maintain consistency between dense checkpoints and MoE layers \(M = \{L_i'\}_{i=N/2}^{N-1}\), input tokens \(\mathbf{X} \in \mathbb{R}^{b \times n \times d}\) (where $b$ represents the batch size, $n$ represents the token length, and $n$ represents the input dimension) are layer normalized inherited from dense checkpoints, resulting in \(\mathbf{X}_{\text{norm}}\). Queries \(\mathbf{Q} \in \mathbb{R}^{b \times (e \times s) \times d}\) are randomly initialized and similarly normalized to align with \(\mathbf{X}_{\text{norm}}\), producing \(\mathbf{Q}_{\text{norm}}\). The layer normalization process ensures the consistency of distributions between the MoE model, input queries, and the pre-trained dense model. The normalized \(\mathbf{X}_{\text{norm}}\) are projected to form keys \(\mathbf{K} \in \mathbb{R}^{b \times n \times d}\) for the cross-attention mechanism. To address numerical instability with randomly initialized queries, \(\mathbf{Q}_{\text{norm}}\) are projected onto a hyperspherical space using L2 normalization. The similarity between \(\mathbf{Q}_{\text{norm}}\) and \(\mathbf{K}\) is computed in this space, yielding similarity logits \(\mathbf{S} \in \mathbb{R}^{b \times (e \times s) \times n}\): $\mathbf{S} = \mathbf{Q}_{\text{norm}} \mathbf{K}^T$. We do not apply L2 normalization to \(\mathbf{K}\) to preserve their scale information, enhancing the matching process. Input slots \(\tilde{\mathbf{X}} \in \mathbb{R}^{b \times (e \times s) \times d}\) for experts are formed by a softmax operation along the $n$ dimension of similarity logits:
\begin{equation}
    \tilde{\mathbf{X}} = \frac{\exp(\mathbf{S}_{ijk})}{\sum_{k'=1}^n \exp(\mathbf{S}_{ijk'})} \mathbf{X}_{\text{norm}}.
\end{equation}
Each expert processes its corresponding input slots \(\tilde{\mathbf{X}_i}\) independently, generating outputs \(\tilde{\mathbf{Y}_i}\). These outputs are then weighted by \(\mathbf{S}\) (after applying a softmax operation along the \((e \times s)\) dimension) to aggregate the experts' contributions, producing the final output \(\mathbf{Y} \in \mathbb{R}^{b \times n \times d}\) of the MoE layer: 
\begin{equation}
    \mathbf{Y} = \frac{\exp(\mathbf{S}_{ijk})}{\sum_{j'=1}^{e \times s} \exp(\mathbf{S}_{ij'k})} \tilde{\mathbf{Y}}.
\end{equation}

In summary, SpheroMoE routing leverages layer normalization, hyperspherical projection, and cross-attention to effectively distribute inputs across experts, ensuring consistency with pre-trained dense models for improved optimization.

\textbf{Expert Regularization}:
To prevent over-specialization and enhance generalization during fine-tuning, we regulate SpheroMoE Routing and expert behavior. We aim to prevent experts from overly focusing on specific inputs and to avoid outputs becoming overly dependent on particular experts.
\underline{For the former}, we introduced a learnable softmax temperature \( T \). During the early stages of fine-tuning, \( T \) is initialized to a large value, causing experts to distribute their attention across all input tokens and preventing early convergence on specific tokens. As training progresses, \( T \) gradually decreases, enabling experts to focus more on specific relevant features and enhancing their specialization where beneficial. Additionally, we added a certain level of expert noise to the similarity logits \( \mathbf{S} \), which improves generalization.
\underline{For the latter}, we utilized stochastic expert dropout, where each expert \( i \) is randomly deactivated with a probability \( p \). It ensures that no single expert becomes a crutch for the entire output, promoting a more balanced utilization of all experts.
These techniques form an expert regularization strategy that maintains expert versatility and mitigates overfitting, ensuring the MoE model performs robustly on downstream datasets.

% \vspace{-4mm}
\textbf{Adaptive Dual-path MoE}:
\label{Dual-path MoE}
To mitigate computational redundancy for less critical tokens and enhance MoE model performance, we introduce the Adaptive Dual-path MoE structure. 
Building on the checkpoint recycling, which enables MoE models to inherit pre-trained knowledge from dense checkpoints and discern crucial from non-crucial tokens, our structure employs two pathways.
The first pathway features fewer core experts with more parameters for processing important tokens. Conversely, the second pathway consists of a larger number of universal experts, each with approximately one-fourth of the parameters of the core experts, designated for handling less critical tokens. The structure is illustrated in Fig.~\ref{fig:dual_path}.
The SpheroMoE Routing mechanism segments input tokens into core and universal slots, assigning them to the respective pathways. The core path processes high-impact tokens, while the universal path handles less important ones, ensuring optimal resource utilization and preserving model accuracy while accelerating the MoE model.

\section{Experiments}
\label{Experiments}
\subsection{Experimental Setups}
\label{experiment setups}
\textbf{Models.} We conduct experiments using Vision Transformer (ViT)~\cite{VIT} and ConvNeXt~\cite{liu2022convnet} to validate our approach. 
Specifically, we transform the ImageNet 21K pre-trained dense checkpoints of ViT-S and ConvNeXt-T into the initialization weights of V-JetMoE-T and C-JetMoE-F through checkpoint recycling. As detailed in Sec.~\ref{Checkpoint Recycling}, V-JetMoE-T comprises dense layers in the first half and is equipped with SpheroMoE layers in the latter half. Each SpheroMoE layer consists of $N/2$ core experts and $N$ universal experts, where $N$ is the number of input tokens. Further details are in Appendix~\ref{model_setting}.

\textbf{Datasets.} We evaluate MoE Jetpack on 8 image classification datasets, including ImageNet-1K~\cite{imagenet}, CIFAR-10, CIFAR-100~\cite{CIFAR}, Flowers~\cite{flower}, Pets~\cite{pets}, STL-10~\cite{STL}, Food-101~\cite{food}, and DTD~\cite{DTD}, encompassing a diverse range of tasks, including object classification, fine-grained species recognition, and texture classification. 

\begin{table}[b]
\vspace{-5mm}
\centering
\small
\caption{Performance comparison on visual recognition tasks with ViT-T and ConvNeXt-F.}
\vspace{2mm}
\begin{adjustbox}{width=1.0\textwidth} % Adjust the box to the desired width
\footnotesize
\begin{tabular}{lr}
\begin{subtable}{0.58\textwidth}
\centering
\addtolength{\tabcolsep}{-1mm}
% \begin{tabular}{@{}l@{\hskip 0.05in}|c@{\hskip 0.05in}c@{\hskip 0.05in}c@{\hskip 0.1in}>{\columncolor{rowunit!15}}c@{\hskip 0.1in}}
\begin{tabular}{c|ccc>{\columncolor{rowunit!15}}c}
\toprule
Dataset ($\downarrow$) & Dense & Dense (21k) & Soft MoE~\cite{softmoe} & MoE Jetpack \\
\midrule
ImgNet-1k & $73.9$ & $75.6$ & \textcolor{black}{$77.1$} & $79.9$~\dplus{$+2.8$} \\
Food-101 & $79.6$ & $86.9$ & \textcolor{black}{$82.0$} & $89.5$~\dplus{$+7.5$} \\
CIFAR-10 & $92.4$ & $97.0$ & \textcolor{black}{$92.9$} & $97.9$~\dplus{$+5.0$} \\
CIFAR-100 & $72.3$ & $81.4$ & \textcolor{black}{$75.9$} & $88.4$~\dplus{$+12.5$} \\
STL-10 & $61.5$ & $83.4$ & \textcolor{black}{$67.7$} & $95.3$~\dplus{$+27.6$} \\
Flowers & $62.4$ & $81.9$ & \textcolor{black}{$70.8$} & $95.4$~\dplus{$+24.6$} \\
Pets & $25.0$ & $68.6$ & \textcolor{black}{$45.5$} & $84.3$~\dplus{$+38.8$} \\
DTD & $49.4$ &$62.5$ & \textcolor{black}{$51.3$} & $69.1$~\dplus{$+17.8$} \\
\bottomrule
\end{tabular}
\caption{ViT-T}
\end{subtable} &

\begin{subtable}{0.44\textwidth}
\centering
\addtolength{\tabcolsep}{-1mm}
\begin{tabular}{ccc>{\columncolor{rowunit!15}}c}
% \begin{tabular}{@{}c@{\hskip 0.05in}c@{\hskip 0.05in}c@{\hskip 0.1in}>{\columncolor{rowunit!15}}c@{}}
\toprule
Dense & Dense (21k) & Soft MoE~\cite{softmoe} & MoE Jetpack \\
\midrule
$76.1$ & $76.4$ & \textcolor{black}{$79.1$} & $80.5$~\dplus{$+1.4$} \\
$86.9$ &$ 89.0$ & \textcolor{black}{$88.7$} & $90.7$~\dplus{$+2.0$} \\
$96.6$ & $97.4$ & \textcolor{black}{$97.3$} & $98.2$~\dplus{$+0.9$} \\
$81.4$ & $84.4$ & \textcolor{black}{$82.8$} & $88.5$~\dplus{$+5.7$} \\
$81.4$ & $92.3$ & \textcolor{black}{$79.4$} & $98.7$~\dplus{$+19.3$} \\
$80.3$ & $94.5$ & \textcolor{black}{$83.3$} & $98.6$~\dplus{$+15.3$} \\
$72.9$ & $87.3$ & \textcolor{black}{$77.4$} & $94.9$~\dplus{$+17.5$} \\
$63.7$ & $68.8$ & \textcolor{black}{$64.7$} & $79.5$~\dplus{$+14.8$} \\
\bottomrule
\end{tabular}
\caption{ConvNeXt-F}
\end{subtable}
\end{tabular}
\label{tab:main_result}
\vspace{-5mm}
\end{adjustbox}
\end{table}

\textbf{Baseline Implementation.} We follow the implementation details outlined by Xu et al.~\cite{xu2024initializing} for comparisons of the dense models. For the MoE models, we employ Soft MoE~\cite{softmoe} as the baseline and have replicated it across all datasets. Our MoE Jetpack and Soft MoE utilize the same training strategies as the dense models to ensure comparison fairness. All implementations were executed using the MMpretrain framework~\cite{2023mmpretrain} on $\text{RTX} 4090$. More information can be found in Appendix~\ref{Training Setting}.

\subsection{Main Results}
Tab.~\ref{tab:main_result} compares the performance of the MoE Jetpack with Dense ViT models (trained from scratch and with pre-trained weights on ImageNet-21k) and Soft MoE models (trained from scratch) on various image classification datasets using ViT-T (a) and ConvNeXt-F (b) architectures. All models maintain approximately the same number of FLOPs.
The MoE Jetpack, which inherits the knowledge from dense checkpoints pre-trained on ImageNet-21k, consistently outperforms MoE models trained from scratch, especially on smaller datasets. These results highlight the effectiveness of MoE Jetpack.

\subsection{Ablations}
We perform ablation studies to assess the impact of various components and hyperparameters within the MoE Jetpack. By default, we use a ViT-T model with the SpheroMoE layer integrated from layers 7 to 12, comprising 98 core experts and 196 universal experts (detailed in Appendix~\ref{model_setting}). The Checkpoint Recycling method transforms dense checkpoints of ViT-S and ViT-T, pre-trained on ImageNet-21k, into initial weights for our V-JetMoE-T model.

\begin{table}[b]
    \vspace{-8mm}
    \centering
    \caption{Ablation Study on MoE Jetpack Components.}
    \vspace{2mm}
    \small
    \setlength{\tabcolsep}{3pt} % Adjust the space between columns
    \begin{adjustbox}{width=0.99\textwidth} % Adjust the box to text width
    \begin{tabular}{c|c|c|ccc|l}
    \toprule
      Soft MoE~\cite{softmoe} & Checkpoints Recycling & SpheroMoE & ImageNet & CIFAR-100 & Flowers  & Mean Acc.\\
    \midrule
    \multicolumn{3}{c|}{Baseline ViT-T} & $73.9$ & $72.3$ & $62.4$ & $69.5$\\
    \midrule
    \checkmark & & & $77.1$ & $75.9$ & $70.8$ & $74.6$~\dplus{$+5.1$}\\
    \checkmark & \checkmark & & $78.4$ & $84.7$ & $91.2$ & $84.8$~\dplus{$+15.3$}\\
     & \checkmark & \checkmark & $79.9$ & $88.4$ & $95.4$ & $\mathbf{87.9}$~\dplus{$+18.4$}\\
    \bottomrule
    \end{tabular}
    \end{adjustbox}
    \label{tab:ablation_1}
    % \vspace{-3mm}
\end{table}

\textbf{Effect of MoE Jetpack Components.} 
We conducted the ablation of two key components of the MoE Jetpack on three datasets. As shown in Tab.~\ref{tab:ablation_1}, integrating Checkpoint Recycling with the Soft MoE baseline significantly improves performance across all datasets, with a mean accuracy increment of $9.8\%$. The SpheroMoE layer further enhances performance, achieving a mean accuracy of $87.9\%$. These results demonstrate the efficacy of both components, especially when used together, highlighting their synergistic effect in boosting performance.

\begin{wraptable}{r}{0.6\textwidth}
\vspace{-4mm}
\centering
\small
\caption{Checkpoint Recycling vs. Sparse Upcycling}
\begin{adjustbox}{width=0.6\textwidth} % Adjust the box to the desired width
\footnotesize
\addtolength{\tabcolsep}{-1mm}
\begin{tabular}{c|c|cc}
    \toprule
    Method & Construction & ImageNet\\
    \midrule
    Sparse Upcycling~\cite{komatsuzaki2022sparseUpcycling} & Copy & $79.1$ \\
    \midrule
    \multirow{4}{*}{\begin{tabular}[c]{@{}c@{}} Checkpoint\\ Recycling \end{tabular}} & Random Sampling & $79.5$\\
    & Uniform Selection & $79.6$\\
    & Graph Partitioning& $79.8$\\
    & Importance-based Sampling& $\mathbf{79.9}$\\
    \bottomrule
\end{tabular}
\end{adjustbox}
\vspace{-3mm}
\label{tab:checkpoint recycle}
\end{wraptable}

% \begin{table}[tbh]

% \centering
% \small
% \caption{\textbf{}}
% % \setlength{\tabcolsep}{3pt} % Adjust the space between columns
% \begin{adjustbox}{width=1\textwidth} % Adjust the box to text width
% % \vspace{2mm}
% \footnotesize
% \begin{tabular}{lr}
% \begin{subtable}{0.68\textwidth}
% \centering
% \small
% \addtolength{\tabcolsep}{-1mm}
%     % \def\arraystretch{1.07}
% \begin{tabular}{c|c|cc}
%     \toprule
%     Method & Construction & ImageNet & Cifar-100 \\
%     \midrule
%     Sparse Upcycling~\cite{komatsuzaki2022sparseUpcycling} & Copy & 79.1 & \\
%     \midrule
%     \multirow{4}{*}{\begin{tabular}[c]{@{}c@{}} Checkpoint\\ Recycling \end{tabular}} & Random Sampling & 79.5 &\\
%     & Uniform Selection & 79.6 &\\
%     & Graph Partitioning& 79.8 &\\
%     & Importance-based Sampling& \textbf{79.9} &\\
%     \bottomrule
% \end{tabular}
% \caption{}
% \label{tab:checkpoint recycle}
% \end{subtable} &

% \begin{subtable}{0.3\textwidth}
% \centering
% \small
% \addtolength{\tabcolsep}{-1mm}
% % \setlength{\tabcolsep}{5.8mm}
% \renewcommand\arraystretch{1.3}
% % \def\arraystretch{1.07}
% \begin{tabular}{cc}
%     \toprule
%     Dense Checkpoint & CIFAR-100 \\
%     \midrule
%     ImageNet-21k~\cite{imagenet} & 1\\
%     DINOv2~\cite{dinov2} & 2\\
%     EVA-02~\cite{EVA02} & 3\\
%     EVA-CLIP~\cite{EVA-CLIP} & 4\\
%     \bottomrule
% \end{tabular}
% \caption{}
% \label{tab:ablation_3}
% \end{subtable}

% \end{tabular}
% \end{adjustbox}
% \label{tab:abla_CheckpointR}
% \end{table}

\textbf{Checkpoint Recycling vs. Sparse Upcycling.}
\label{ablation Checkpoint Recycling}
To compare the four checkpoint recycling strategies mentioned in Sec.~\ref{Checkpoint Recycling} and the method of using duplicated MLPs to construct experts in Sparse Upcycling~\cite{komatsuzaki2022sparseUpcycling}, we conducted experiments on ImageNet. For fairness, we also employed our SpheroMoE layer in the Sparse Upcycling. The results, summarized in Tab.~\ref{tab:checkpoint recycle}, show that Importance-Based Sampling achieves the highest performance, demonstrating its effectiveness in leveraging critical weights to enhance model performance and convergence speed. Additionally, Checkpoint Recycling is highly flexible, allowing the construction of experts of varying sizes to meet different needs, a feature not provided by sparse upcycling.

\begin{wrapfigure}{rb}{0.55\textwidth}
\vspace{-8mm}
\begin{center}
   \includegraphics[width=1.0\linewidth]{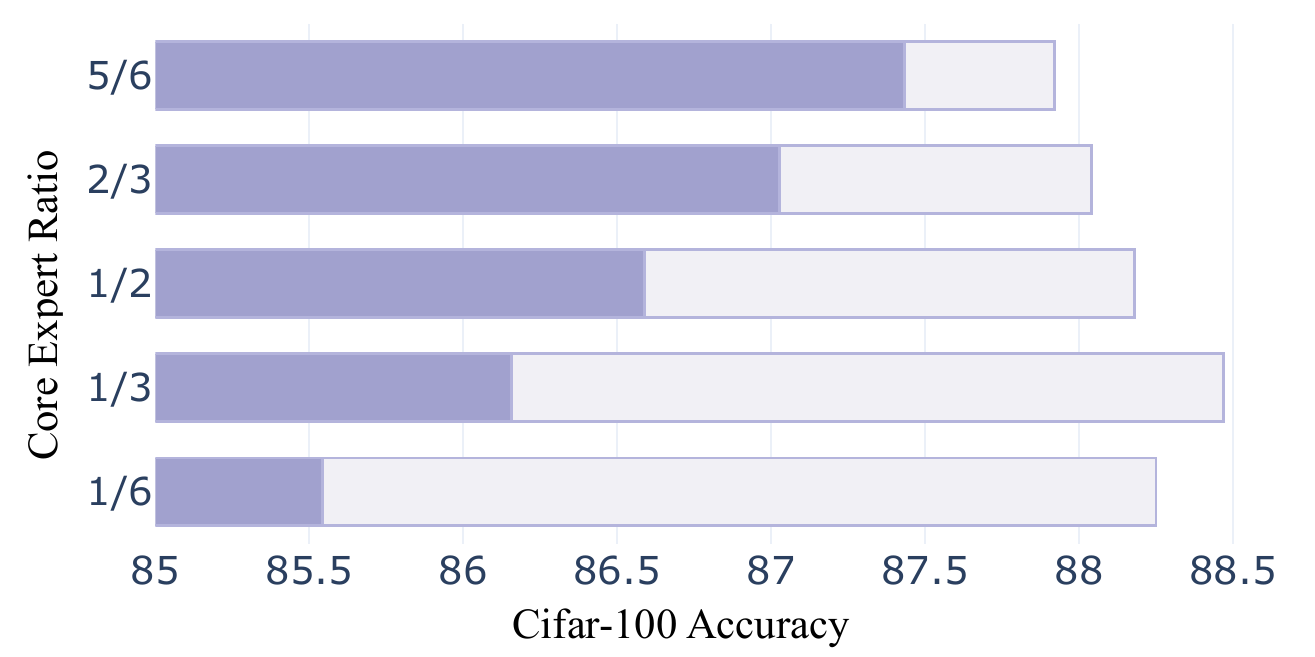}
\end{center}
    \vspace{-3mm}
   \caption{
   This chart shows CIFAR-100 accuracy across different ratios of core (dark) to universal (light) experts, highlighting optimal performance at a 1/3 core ratio.
   }
\label{fig:core expert ratio}
\vspace{-6mm}
\end{wrapfigure}

\textbf{Core Experts Ratio.} 
To assess the impact of the Adaptive Dual-path MoE structure introduced in Sec.~\ref{SpheroMoE Layer} on the accuracy of MoE models, we aimed to determine the ideal balance between performance and resource allocation. We conducted experiments on the Cifar-100 dataset with a constant number of total experts, varying the ratio of core experts. The results, illustrated in Fig.~\ref{fig:core expert ratio}, indicate that optimal accuracy is achieved when the proportion of core experts is set at $1/3$.

\textbf{Different MoE Jetpack Configurations.} 
This part evaluates the impact of various MoE Jetpack configurations on model performance, as summarized in Tab.~\ref{tab:ablation_modelsize}. The experiments focus on the placement of SpheroMoE layers, the number of experts per layer, and the base size of converted dense checkpoints. Results indicate that more SpheroMoE layers generally enhance performance, though placing it before layer $7$ slightly hurt the performance. Consequently, SpheroMoE layers were incorporated into layers $7{-}12$. Additionally, models with more experts exhibit improved accuracy, highlighting the benefits of increased expert specialization and diversity. Models converted from larger dense checkpoints demonstrate superior performance. These findings suggest that MoE network performance can be improved by increasing the number of MoE layers, incorporating more experts, and utilizing larger base models.
\begin{table}[t]
    \centering
    \small
    \vspace{-5mm}
    \caption{Comparison of Model Variants with Different Configurations}
    \vspace{2mm}
    \setlength{\tabcolsep}{5pt} % Adjust the space between columns
    \begin{adjustbox}{width=.99\textwidth} % Adjust the box to text width
    \begin{tabular}{c|ccccc|cc}
    \toprule
    model & Weight Init. & MoE Layers & Expert Number & Param (M) & FLOPs (G)& CIFAR-100 & ImageNet \\
    \midrule
     ViT-T & - & - & - & $6$ & $1.1$ & $72.3$ & $73.9$ \\
     Soft MoE-T~\cite{softmoe} & - & $7{:}12$ & $197$ & $354$ & $1.2$ & $75.9$ & $77.1$ \\
     Soft MoE-S~\cite{softmoe} & - & $7{:}12$ & $197$ & $1412$ & $4.5$ & $77.5$ & $80.3$ \\
    \midrule
     ViT-T & \checkmark & - & - & $6$ & $1.1$ & $81.4$ & $75.5$ \\
     V-JetMoE-T & \checkmark & $11{:}12$ & core: $98$, univ: $196$ & $92$ & $1.1$ & $87.4$ & - \\
     V-JetMoE-T & \checkmark & $9{:}12$ & core: $98$, univ: $196$ & $179$ & $1.1$ & $87.8$ & - \\
     V-JetMoE-T & \checkmark & $5{:}12$ & core: $98$, univ: $196$ & $352$ & $1.2$ & $86.7$ & - \\
    \midrule
     V-JetMoE-T & \checkmark & $7{:}12$ & core: $32$, univ: $64$ & $89$ & $0.8$ & $87.8$ & - \\
     V-JetMoE-T & \checkmark & $7{:}12$ & core: $64$, univ: $128$ & $175$ & $1.0$ & $88.0$ & - \\
    \midrule
     V-JetMoE-T & \checkmark & $7{:}12$ & core: $98$, univ: $196$ & $265$ & $1.1$ & \underline{$88.4$} & \underline{$79.9$} \\
     V-JetMoE-S & \checkmark & $7{:}12$ & core: $98$, univ: $196$ & $1058$ & $4.3$ & $\mathbf{89.9}$ & $\mathbf{82.4}$ \\
    \bottomrule
    \end{tabular}
    \end{adjustbox}
    \vspace{-5mm}
    \label{tab:ablation_modelsize}
\end{table}

% \begin{table}[tbh]
%     \centering
%     \small
%     % \vspace{-10mm}
%     \caption{Comparison of Model Variants with Different Configurations}
%     \vspace{2mm}
%     \setlength{\tabcolsep}{3pt} % Adjust the space between columns
%     \begin{adjustbox}{width=.98\textwidth} % Adjust the box to text width
%     \begin{tabular}{c|ccccc|cc}
%     \toprule
%     model & Weight Init. & MoE Layers & Expert number & Param(M) & FLOPs & CIFAR-100 & ImageNet \\
%     \midrule
%      ViT-T & - & - & - & 6 & 1.1 & 72.3 & 73.9 \\
%      Soft MoE-T~\cite{softmoe} & - & 7:12 & 197 & 354 & 1.2 & 75.9 & 77.1 \\
%      Soft MoE-S~\cite{softmoe}& - & 7:12 & 197 & 1412 & 4.5 & 77.5 & 80.3 \\
%     \midrule
%      ViT-T & \checkmark & - & - & 6 & 1.1 & 81.4 & 75.5 \\
%      V-JetMoE-T & \checkmark & 11:12 & $[core:98, univ:196]$ & 92 & 1.1 & 87.4 & - \\
%      V-JetMoE-T & \checkmark & 9:12 & $[core:98, univ:196]$ & 179 & 1.1 & 87.8 & - \\
%      V-JetMoE-T & \checkmark & 5:12 & $[core:98, univ:196]$ & 352 & 1.2 & 86.7 & - \\
%      \midrule
%      V-JetMoE-T & \checkmark & 7:12 & $[core:32, univ:64]$ & 89 & 0.8 & 87.8 & - \\
%      V-JetMoE-T & \checkmark & 7:12 & $[core:64, univ:128]$ & 175 & 1.0 & 88.0 & - \\
%      \midrule
%      V-JetMoE-T & \checkmark & 7:12 & $[core:98, univ:196]$ & 265 & 1.1 & \underline{88.4} & \underline{79.9} \\
%      V-JetMoE-S & \checkmark & 7:12 & $[core:98, univ:196]$ & 1058 & 4.3 & \textbf{89.9} & \textbf{82.4} \\
%     \bottomrule
%     \end{tabular}
%     \end{adjustbox}
%     \vspace{-4mm}
%     \label{tab:ablation_modelsize}
% \end{table}
% 

\vspace{-2mm}

\subsection{Analysis}
In this section, we investigate the influence of the MoE Jetpack on enhancing the convergence speed of MoE models when trained on the ImageNet and CIFAR-100 datasets. Additionally, we provide some intuition regarding the attention patterns of the experts and the contribution of each expert to the final results.

\begin{figure}[b]
    \vspace{-5mm}
    \centering
    \includegraphics[width=1.\linewidth]{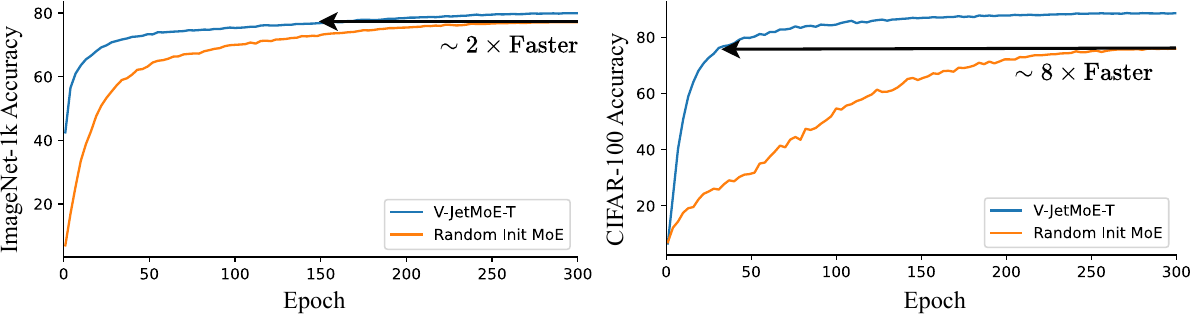}
    \vspace{-2mm}
    \caption{Comparison of convergence speeds using MoE Jetpack versus training from scratch on ImageNet (left) and CIFAR-100 (right). MoE Jetpack achieves target accuracies significantly faster, demonstrating a 2x speed increase on ImageNet and an 8x increase on CIFAR-100.}
    % \vspace{-5mm}
    \label{fig:acc}
\end{figure}

\textbf{Accelerating MoE Convergence with MoE Jetpack.}
The impact of MoE Jetpack on convergence speed is evident in Fig.~\ref{fig:acc} for ImageNet (left) and CIFAR-100 (right). In both cases, models with MoE Jetpack reach the target accuracy significantly faster. For ImageNet, the model with MoE Jetpack achieves approximately $77\%$ top.1 accuracy before $150$ epochs, \textbf{$\mathbf{2}$ times faster} than training from scratch. Notably, for smaller datasets like CIFAR-100, the acceleration effect of MoE Jetpack is more pronounced: The model with MoE Jetpack reaches the $76\%$ top.1 accuracy at around $40$ epochs, \textbf{$\mathbf{8}$ times faster} than the model without it. These results demonstrate that MoE Jetpack substantially accelerates convergence speed, enhancing fine-tuning efficiency and reducing computational resources.

\begin{figure}[tb]
    \vspace{-5mm}
    \centering
    \includegraphics[width=1.\linewidth]{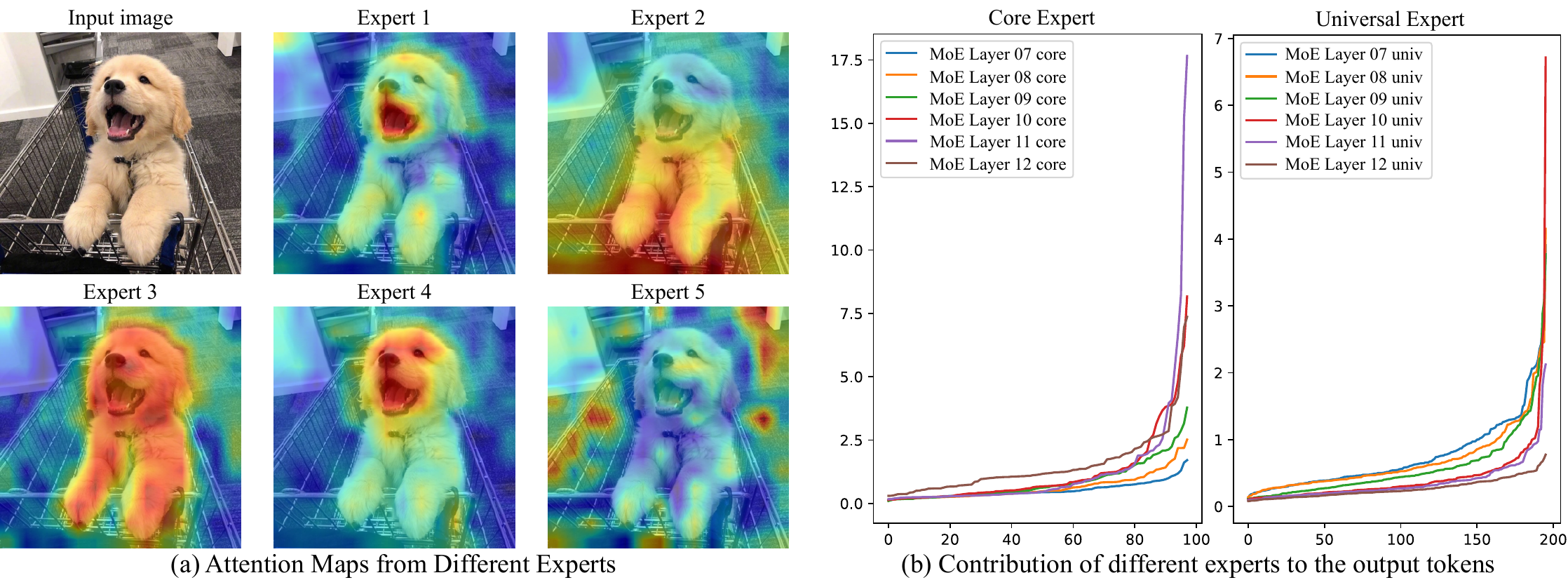}
    \caption{
    (a) This figure illustrates the attention maps generated by five experts in response to an input image, highlighting the experts' specialization.
    (b) These line charts show varying contributions of core and universal experts, with core experts' influence peaking in later layers, emphasizing their detailed feature refinement, contrasted with the consistent input of universal experts.
    }
    \vspace{-3mm}
    \label{fig:vis}
\end{figure}

\textbf{Intuition of Expert Attention Patterns.}
We visualize the attention maps of experts in Fig.~\ref{fig:vis}(a), which illustrates that different experts focus on different parts of the input image. This diversity in attention suggests that each expert specializes in capturing unique aspects of the input, enhancing the model's ability to represent features comprehensively. The specialization allows the MoE model to combine multiple perspectives, resulting in a more robust and detailed understanding of the input.

\textbf{Contribution of Each Expert to Final Results.}
Fig.~\ref{fig:vis}(b) demonstrates the varying contributions of core and universal experts across different layers of the MoE model. Core experts show an increasing influence in the later layers, emphasizing their role in refining specific and highly relevant features. Additionally, the contributions among core experts are markedly uneven, some experts can impact output tokens $17\times$ more than others, reflecting greater specialization and diversity in their focus areas. In contrast, universal experts maintain a relatively consistent contribution level, indicating a more uniform integration of broader contextual information throughout the network. This hierarchical structure, balancing the specialized refinement by core experts with the generalized understanding provided by universal experts, enhances the model's overall performance and robustness.

% \textbf{Visualization of the diversity of experts.}
% In this section, we visualize the diversity among experts. Figure~\ref{fig:vis}(a) shows the image regions attended to by several experts in the first layer of the Mixture of Experts (MoE) model, highlighting their distinct focus areas. Figure~\ref{fig:vis}(b) presents the summed weights of the experts' contributions to the output combined tokens. Due to the Adaptive Dual-path MoE method, we separately analyze two types of experts. The results indicate that certain experts significantly impact the output, with core experts generally exerting greater influence than univ experts.

%In this subsection, we visualize the diversity among experts. As shown in Figure~\ref{fig:vis}(a), we illustrate the image regions attended to by several experts in the first layer of the MoE model. It is evident that different experts focus on different areas. In Figure~\ref{fig:vis}(b), we sum the weights assigned to the outputs of the experts to show their contribution to the combined tokens. Specifically, we perform separate statistics for the two types of experts due to using the Adaptive Dual-path MoE method. It can be observed that there are some experts that contribute significantly to the output, with the univ experts generally having a greater impact compared to the core experts.

\section{Related Work}
\label{Related Work}
\vspace{-2mm}
\textbf{Sparsely activated Mixture of Experts (MoE).} 
Scaling Laws~\cite{scalinglaw} indicate that increasing model parameters can enhance performance. However, traditional densely activated models (dense models)~\cite{VIT,liu2022convnet} activate all parameters for every input, resulting in high computational costs as models scale. In contrast, MoE models~\cite{deepseekmoe, lin2024moe, jiang2024mixtral, wu2023mole} activate only a subset of parameters for specific input tokens, enabling efficient scaling to trillions of parameters with sublinear increases in computational costs~\cite{lepikhin2020gshard, riquelme2021scaling, fedus2022switch}. To optimize input token allocation among experts, various routing mechanisms have been developed. BASELayer~\cite{lewis2021base} formulates token-to-expert allocation as a linear assignment problem, while EC-CF2~\cite{zhou2022mixture} propose expert choice routing, soft routing methods like SMEAR~\cite{muqeeth2023soft}, and Soft MoE~\cite{softmoe} implicit soft assignments involving all tokens. However, few studies explore leveraging dense model checkpoints to accelerate MoE training~\cite{komatsuzaki2022sparseUpcycling}.

\textbf{Knowledge transfer with pre-trained models.} 
Knowledge transfer occurs between \textbf{identical} or \textbf{distinct} models. Pre-training followed by fine-tuning is well-established for \textbf{identical models}, utilizing large datasets through supervised learning (e.g., ImageNet21k~\cite{imagenet}, JFT-300M~\cite{JFT300M}) or self-supervised methods (e.g., BERT~\cite{BERT}, CLIP~\cite{clip}, MAE~\cite{MAE}, DINO~\cite{dinov2}, EVA~\cite{EVA02, EVA-CLIP}). These approaches produce foundation models with broad applicability, and subsequent fine-tuning consistently improves performance.
For \textbf{distinct models}, knowledge distillation~\cite{knowledgedistillation} trains a smaller student model to mimic the larger teacher model, enhancing efficiency. Additional strategies include weight pruning~\cite{weightpruning, ashkboos2023slicegpt, ashkboos2023slicegpt, xia2023sheared, yu2022width}, which removes redundant parameters, and weight selection~\cite{xu2024initializing} initializes a smaller model with a subset of weights from a pre-trained larger model.

\textit{Research on transferring knowledge from dense checkpoints to MoE models is limited}. MoEfication~\cite{zhang2022moefication} partitions a dense model into MoE components, while Sparse Upcycling~\cite{komatsuzaki2022sparseUpcycling} replicates a dense model multiple times to form a MoE model. Our MoE Jetpack recycles important weights from larger dense checkpoints to initialize experts of various sizes, combining the flexibility of knowledge transfer across distinct models with the efficiency of transfer between identical models.

\vspace{-3mm}
\section{Conclusion}
\label{Conclusion}
\vspace{-3mm}
% In this paper, we introduced MoE Jetpack, a novel framework for fine-tuning pre-trained dense checkpoints into mixture of experts (MoE) models. Our approach leverages checkpoint recycling, and the hyperspherical adaptive MoE (SpheroMoE) layer to enhance convergence speed and model accuracy. The MoE Jetpack demonstrated significant improvements across various visual tasks while maintaining computational efficiency.
In this paper, we introduced MoE Jetpack, a novel framework for fine-tuning pre-trained dense checkpoints into Mixture of Experts. Our approach leverages checkpoint recycling, which inherits the knowledge of open-source dense checkpoints and the hyperspherical adaptive MoE (SpheroMoE) layer to enhance fine-tuning performance. These innovations contribute to improved convergence speed and model accuracy. The MoE Jetpack significantly improved various visual tasks while maintaining computational efficiency.

The \textbf{limitation} of our approach is its dependency on the quality of pre-trained dense checkpoints; poorly trained or inadequately generalized dense models could limit the performance enhancements. Additionally, while our experiments focused on visual tasks, further research is needed to validate the generalizability of MoE Jetpack across other domains, such as natural language processing and reinforcement learning. We believe future work will address these limitations, enhance the scalability and robustness of the framework, and extend MoE applicability to a broader range of tasks.

%%%%%%%%%%%%%%%%%%%%%%%%%%%%%%%%%%%%%%%%%%%%%%%%%%%%%%%%%%%%
\bibliographystyle{IEEEtran}     
\bibliography{references.bib}

%%%%%%%%%%%%%%%%%%%%%%%%%%%%%%%%%%%%%%%%%%%%%%%%%%%%%%%%%%%%
\newpage
\appendix

% \section{Appendix / supplemental material}
\section{Detailed Model Configurations}
\label{model_setting}
In this section, we present the detailed model configurations for the main experiments in Sec.~\ref{Experiments} in  Tab.~\ref{tab:config}. We refer to pre-trained dense checkpoints as \textbf{predecessors} and the derived MoE models as \textbf{successors}. We use ImageNet-21k pre-trained predecessor from timm with our Checkpoint Recycling algorithm to generate initialized weights for the successor.

\vspace{-5mm}
\begin{table}[tbh]
    \centering
    \small
    \caption{Configurations for Models.}
    \vspace{2mm}
    \addtolength{\tabcolsep}{-1mm}
    \begin{tabular}{c|cc|cc}
    \toprule
    Configuration & \multicolumn{2}{c|}{Successors} & \multicolumn{2}{c}{Predecessors}\\
    \midrule
    Model & V-JetMoE-T & C-JetMoE-F & ViT-S/16 & ConvNext-T\\
    FLOPs (G) & $1.1$ & $1.1$ & $1.1$ & $1.1$ \\
    Initialization & Checkpoint Recycling & Checkpoint Recycling & ImageNet-21k & ImageNet-21k  \\
    MoE Layers & $7$:$12$ & $10$:$18$ & - & - \\
    % MoE Type & SpheroMoE & SpheroMoE & - & - \\
    Core Expert Number & $98$ & [$98$, $24$] & - & - \\
    Universal Expert Number & $196$ & [$196$, $48$] & - & - \\
    
    % Parameters Number & 265 & ? & ? & ? \\

    \bottomrule
    \end{tabular}
    \label{tab:config}
    \vspace{-5mm}
\end{table}

\section{Experiment Settings and Time Costs}
\label{Training Setting}
In this section of the appendix, we provide a comprehensive description of the training settings used in our experiments.
Tab.~\ref{tab:basic_recipe} outlines the standard training configuration utilized across our experiments. Tab.~\ref{tab:training_settings} details the dataset-specific training configurations, capturing variations in batch size, warmup epochs, total training epochs, and drop path rates for each dataset employed in our experiments. \par
Our experiments were conducted on RTX 4090 GPU. Training V-JetMoE-T on the CIFAR-100 dataset (60,000 images) required 2.5 GPU hours while training on the ImageNet-1K dataset (1,281,167 images) required 120 GPU hours. Training C-JetMoE-F on CIFAR-100 also required 2.5 GPU hours and 156 GPU hours on ImageNet-1K. For V-JetMoE-S, training on CIFAR-100 required 8 GPU hours and 200 GPU hours on ImageNet-1K. Compared to the original dense models (ViT-Tiny, ConvNeXt-Femto, ViT-Small), our method achieves nearly equivalent training times.\par
For all the experiments presented in our paper, we required $3,300$ GPU hours for training. In total, we spent approximately $8,000$ GPU hours for exploration and validation of our work.

\begin{table}[tbh]
    \centering
    
    \begin{minipage}{0.45\linewidth}
        \centering
        \caption{Our basic recipe for model training.}
        \vspace{2mm}
        \addtolength{\tabcolsep}{-2mm}
            \begin{tabular}{l c}
        \toprule
        Training Setting & Configuration \\ 
        \midrule
        image resolution & $224 \times 224$\\
        optimizer & AdamW\cite{adamw} \\ 
        base learning rate & $4 \times 10^{-3}$ \\ 
        weight decay & $0.05$ \\ 
        optimizer momentum & $\beta_1, \beta_2=0.9, 0.999$ \\ 
        batch size & $4096$ \\ 
        training epochs & $300$ \\ 
        learning rate schedule & cosine decay \\ 
        warmup epochs & $50$ \\ 
        warmup schedule & linear \\ 
        % layer-wise lr decay~\cite{lwld1,lwld2}  & None \\ 
        randaugment~\cite{cubuk2020randaugment}  & $(9, 0.5)$ \\ 
        mixup~\cite{zhang2018mixup}  & $0.8$ \\ 
        cutmix~\cite{yun2019cutmix}  & $1.0$ \\ 
        random erasing~\cite{zhong2020random}  & $0.25$ \\ 
        label smoothing~\cite{labelsmooth}  & $0.1$ \\ 
        layer scale~\cite{layerscale}  & $1 \times 10^{-6}$ \\ 
        \bottomrule
    \end{tabular}
    \label{tab:basic_recipe}
    \end{minipage}
    \hfill
    \begin{minipage}{0.45\linewidth}
        \centering
        \caption{Hyper-parameter setting on ViT-T.}
        \vspace{2mm}
        \addtolength{\tabcolsep}{-1.5mm}
        \begin{tabular}{lcccc}
            \toprule
            Setting   & \makecell{Batch \\ Size} & \makecell{Warmup \\ Epochs} & \makecell{Training \\ Epochs} & \makecell{Drop \\ Path Rate} \\ 
            \midrule
            C-10    & $512$ & $50$ & $300$ & $0.1$ \\ 
            C-100   & $512$ & $50$ & $300$ & $0.1$ \\ 
            Pets    & $512$  & $100$ & $600$ & $0.1$ \\ 
            Flowers & $512$  & $100$ & $600$ & $0.1$ \\ 
            STL-10  & $512$  & $50$ & $300$ & $0$ \\ 
            Food101 & $512$ & $50$ & $300$ & $0.1$ \\ 
            DTD     & $512$  & $100$ & $600$ & $0.2$ \\ 
            IN1k    & $4096$ & $50$ & $300$ & $0$ \\ 
            \bottomrule
        \end{tabular}
        \label{tab:training_settings}
    \end{minipage}
\end{table}

\newpage
\section{Implementation of SpheroMoE Layer}
\label{pseudo code}
\begin{lstlisting}[caption={Simple implementation of SpheroMoE.}, label={lst:spheromoe_layer}]
def parallel_expert_forward(x, experts)
    """
    Traditional MoE models use a for loop to process each token through the experts. 
    
    By merging all expert weights into a large matrix, our implementation allows for a single matrix multiplication operation for each layer across all tokens and experts, replacing multiple individual operations.
    """
    x = einsum(x, experts.weight_1, "b e s d1, e d2 d1 -> b e s d2")
    x = x + rearrange(experts.bias_1, "e d2 -> () e () d2")
    x = experts.act(x)
    x = einsum(x, experts.weight_2, "b e s d2, e d1 d2 -> b e s d1")
    x = x + rearrange(experts.bias_2, "e d1 -> () e () d1")
    return x
        
def spheromoe_layer(X, Q, T, core_experts, univ_experts):
    """
    Performs the Spheromoe layer operation.
    
    Parameters:
    X (tensor): tensor with shape (batch, token_num, channel).
    Q (tensor): tensor with shape (expert_num, slots_per_expert, channel).
    T (float): temperature parameter for the softmax function.
    core_experts, univ_experts (expert): expert weight for MoE layer.
    
    Returns:
    tensor: Output tensor after applying the Spheromoe layer operations.
    """
    X_norm = inherit_layer_norm(X, dim=-1)
    Q_norm = l2_norm(inherit_layer_norm(Q, dim=-1))
    K = K_project(X_norm)
    
    # Compute similarity logits S.
    S = einsum(K, Q_norm, "b n d, e s d -> b n e s")

    # Add normal noise
    noise = normal_noise(S) * self.noise_mult
    S = S + noise
    
    # Apply softmax to similarity logits.
    Dispatch = softmax(S/T, dim=1)
    Combine = softmax(S/T, dim=[-1,-2])
    
    # Token dispatch.
    X_hat = einsum(Dispatch, X_norm, "b n d, b n e s -> b e s d")
    X_core = X_hat[:, :core_num, :, :]
    X_univ = X_hat[:, core_num:, :, :]
    
    # Using core experts and universal experts processes each slot.
    Y_hat = stack([
        parallel_expert_forward(X_core, core_experts),
        parallel_expert_forward(X_univ, univ_experts)
    ], dim=1)

    # Expert dropout.
    Y_hat = expert_drop(Y_hat)
    
    # Token combine.
    Y = einsum(Combine, Y_hat, "b n e s, b e s d -> b n d" )
    
    return Y
\end{lstlisting}

\section{Dynamic Allocation and Focus Regions of Experts in MoE Jetpack}
\label{append_vis}
\begin{figure}[h]
    \centering
    \includegraphics[width=1.\linewidth]{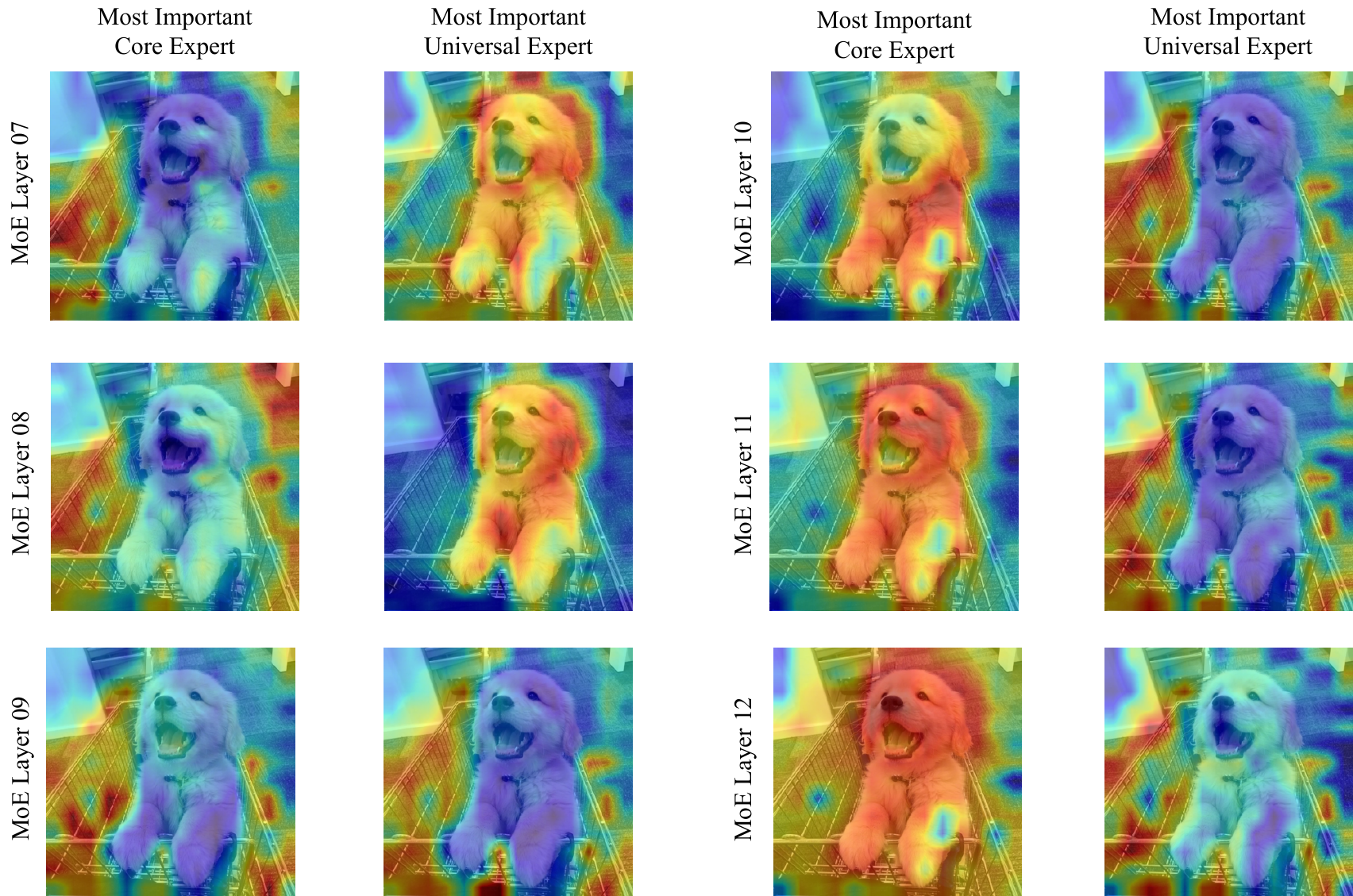}
    \caption{Visualization of the attention map identified by the most important core experts and universal experts across different layers (MoE Layer 07 to  MoE Layer 12). The images show the regions deemed most relevant by each type of expert at each layer.}
    \vspace{-6mm}
    \label{fig:append_vis}
\end{figure}

In this section, we discuss the dynamic allocation and focus regions of core and universal experts across different layers of MoE Jetpack. We used the same test images as in the main text, visualizing the focus regions of the most important (i.e., those with the highest output contribution) core and universal experts for each MoE layer in Fig.~\ref{fig:append_vis}. The corresponding contribution values for these experts are listed in Tab.~\ref{table:expert_contributions}.

Our findings are as follows: Initially, in the shallower network layers (MoE Layer 7 and 8), the core experts contribute less than the universal experts, and their focus regions are relatively dispersed. As the network deepens, in MoE Layer 9, the most important core and universal experts show similar contribution values and focus regions. With further depth (MoE Layers 10, 11, and 12), the dominance of the core experts becomes increasingly evident, with significantly higher contribution values than the universal experts. Core experts focus on prominent objects in the images and are inclined to capture global information.

These experts' dynamic allocation and different focus region tendencies are crucial to our method. Different experts have varying capabilities in extracting information at various granularities, and the network facilitates collaboration among these experts to produce the final output. This illustrates the effective utilization of expert diversity in the MoE model.

\begin{table}[h]
\centering
\caption{Contribution values of core and universal experts across network layers.}
\vspace{2mm}
\begin{tabular}{c|c|c}
\toprule
MoE Layer & Core Expert Contribution & Universal Expert Contribution \\ 
\midrule
7 & $1.71$ & $3.91$ \\
8 & $2.52$ & $4.16$ \\
9 & $3.78$ & $3.77$ \\
10 & $8.17$ & $6.71$ \\
11 & $17.66$ & $2.12$ \\
12 & $7.36$ & $0.77$ \\
\bottomrule
\end{tabular}
\label{table:expert_contributions}
\end{table}

\section{Broader Impacts}
\label{broader impacts}
The proposed MoE Jetpack framework significantly enhances the accessibility and efficiency of MoE models by utilizing pre-existing dense checkpoints to substantially reduce the computational costs associated with training these models from scratch. This method not only minimizes the environmental footprint by decreasing the reliance on extensive GPU resources but also bridges the resource gap, facilitating wider adoption and fostering innovation across the AI community. Additionally, our commitment to open-sourcing all experimental code promotes greater transparency and collaboration in research. We have carefully considered the potential societal impacts of our method and believe it does not pose any significant ethical or fairness concerns, thereby ensuring its responsible application.

% The proposed MoE Jetpack framework significantly facilitates the application of MoE models by enabling effective fine-tuning of dense checkpoints into MoE models. This approach can reduce the environmental impact associated with extensive GPU usage for training large models from scratch. This approach fills the resource gap, promoting broader adoption and innovation within the AI community. Furthermore, our commitment to fully open-sourcing all experimental code enhances research transparency and collaboration. To the best of our knowledge, our method does not pose any potential negative societal impacts, ensuring its ethical and fair application.

% Optionally include supplemental material (complete proofs, additional experiments and plots) in appendix.
% All such materials \textbf{SHOULD be included in the main submission.}

%%%%%%%%%%%%%%%%%%%%%%%%%%%%%%%%%%%%%%%%%%%%%%%%%%%%%%%%%%%%

\end{document}